\documentclass{article}
\usepackage{spconf,amsmath,graphicx}

\usepackage{framed,multirow}
\usepackage{amssymb}
\usepackage{latexsym}
\usepackage{lipsum}
\usepackage{textcomp}
\usepackage{url,cite}
\usepackage[table]{xcolor}
\usepackage{graphicx}
\usepackage[tight,footnotesize]{subfigure}
\usepackage{dsfont}
\usepackage{tabularx, tabulary, multirow, bigstrut}
\usepackage{booktabs}
\usepackage[american]{babel}
\usepackage[vlined]{algorithm2e} 
\usepackage{fixltx2e}
\usepackage[pagebackref=false,breaklinks=true,letterpaper=true,colorlinks,bookmarks=false]{hyperref}
\graphicspath{ {./figs/} }

\DeclareMathOperator*{\argmax}{\mathrm{argmax}}

\title{Long and Short Memory Balancing in Visual Co-Tracking \\using Q-learning}
\name{Kourosh Meshgi$^{\star}$ \qquad Maryam Sadat Mirzaei$^{\star\dagger}$ \qquad Shigeyuki Oba$^{\star}$\thanks{This article is based on results obtained from a project commissioned by the NEDO and was supported by Post-K application development for exploratory challenges from Japan's MEXT.}}
\address{$^{\star}$ Graduate School of Informatics, Kyoto University, Japan \\
    $^{\dagger}$ RIKEN Center for Advanced Intelligence Project (AIP), Japan}

\begin{document}
\setlength{\abovedisplayskip}{3pt}
\setlength{\belowdisplayskip}{3pt}
%
\maketitle
\begin{abstract}
Employing one or more additional classifiers to break the self-learning loop in tracing-by-detection has gained considerable attention. Most of such trackers merely utilize the redundancy to address the accumulating label error in the tracking loop, and suffer from high computational complexity as well as tracking challenges that may interrupt all classifiers (e.g. temporal occlusions). We propose the active co-tracking framework, in which the main classifier of the tracker labels samples of video sequence, and only consults auxiliary classifier when it is uncertain. Based on the source of the uncertainty and the differences of two classifiers (e.g. accuracy, speed, update frequency, etc.), different policies should be taken to exchange the information between two classifiers. Here, we introduce a reinforcement learning approach to find the appropriate policy by considering the state of the tracker in a specific sequence. The proposed method yields promising results in comparison to the best tracking-by-detection approaches.
\end{abstract}
\section{Introduction}
\label{sec1}
Tracking-by-detection methods are built around the idea that a single classifier separates the target from its background by labeling (or filtering) several samples from the input image, labeling them, and extrapolating these samples to estimate the current target location and size. This classifier needs to be updated to cope with recent target transformations as well as other challenging factors such as changes in illumination, camera pose, cluttered background, and occlusions. The update process is mainly done using the labels that the classifier selected for the samples, in a self-supervised learning fashion.

A classifier is not always certain about the output labels. Whether it is inefficient features for certain input images, insufficient model complexity to separate some of the samples, lack of proper training data, missing information in the input data (e.g., due to occlusion), or technically speaking, having an input sample that falls very close to decision boundary of the classifier, hampers the classifier ability to be sure about its label and increase the risk of misclassification. Especially in the case of online learning, novel appearances of the target, background distractors, and non-stationarity of the label distribution\footnote{A sample might be considered as foreground but later the label become obsolete or become a part of the background.} promotes the uncertainty of the classifier.

\begin{figure}[!t]
\centering
\includegraphics[width=0.8\linewidth]{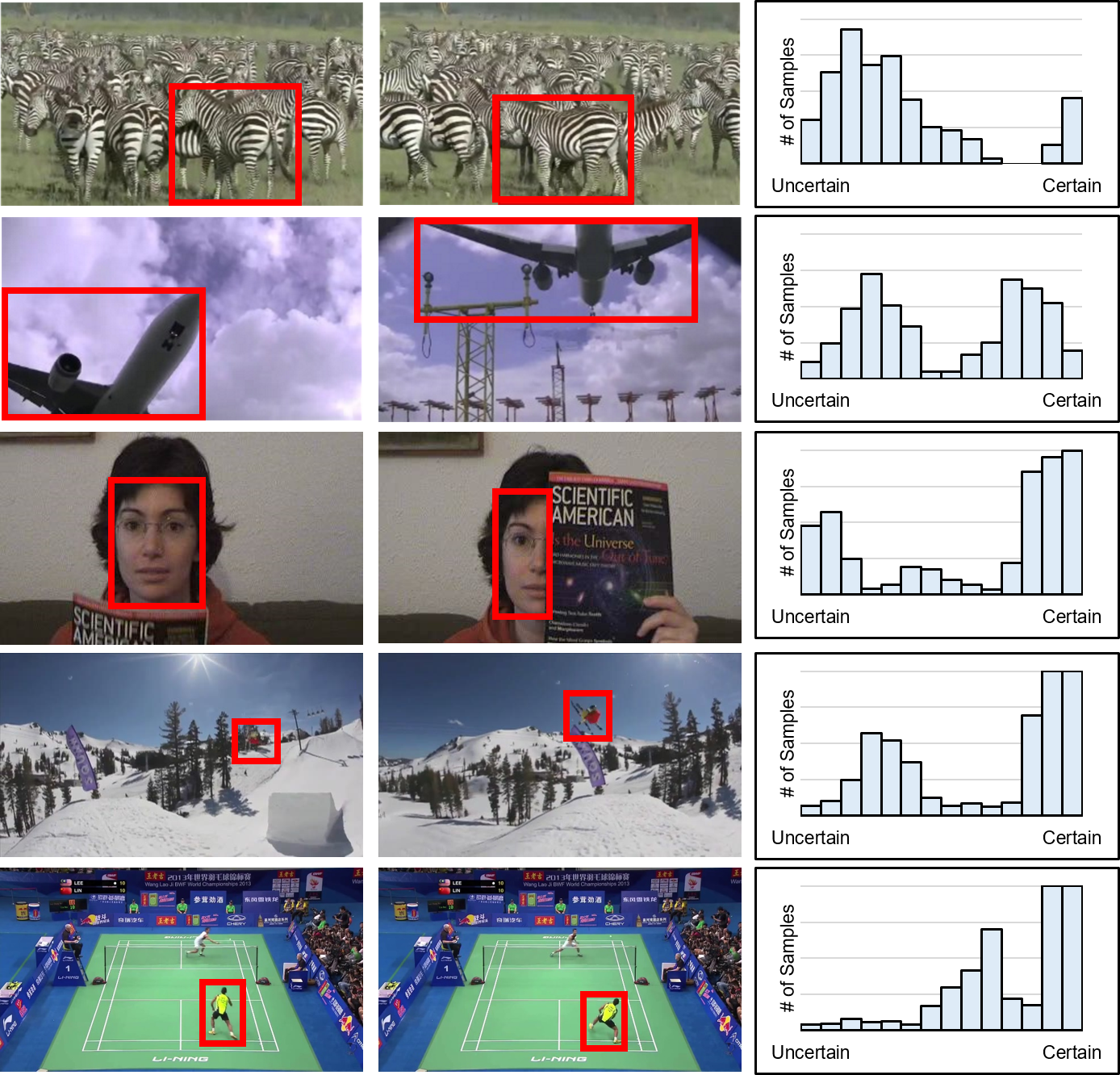}
\caption{Consider a classifier of tracking-by-detection that uses color and shape features and is trained on video frames leading to the frame on the left column. When classifying $n_s$ samples from the frame in the middle column, the uncertainty for all samples may have different trends, as plotted in the uncertainty histogram in right panels. The histogram may be skewed toward certainty, uncertainty (e.g. due to feature failures or occlusion), bimodal (where usually background is easy to separate but the foreground is ambiguous), etc. In co-tracking frameworks, various patterns of uncertainty require different policies to enhance tracking performance.}
\label{fig:concept}
\vspace{-0.5 cm}
\end{figure}

\begin{figure*}
\centering
\includegraphics[width=0.8\linewidth]{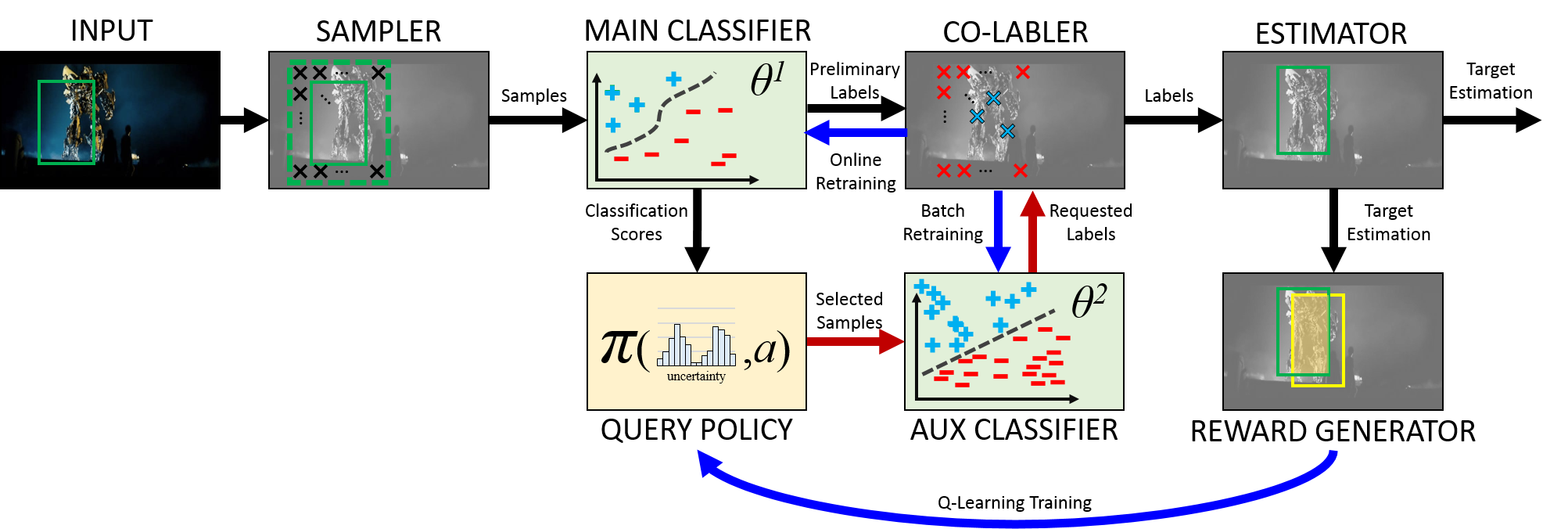}
\caption{Schematic of the system. The proposed tracker, collect samples from a pre-defined area around the last known target location. The classification scores are fed to query the policy unit which selects the best value for uncertainty margin hyperparameter. If needed, the aux classifier is queried for the label of the samples. The classifiers are then updated using co-labeled samples and the target state is estimated. In the training phase, the target estimation is compared to the ground truth and their normalized intersection is used as a reward to train the query policy Q-table.}
\label{fig:schematic}
\vspace{-0.5 cm}
\end{figure*}
Furthermore, the self-supervised learning loop may lead to model drift due to the accumulation of label errors, and many studies have tried to tackle this problem by using robust loss functions for the classifier \cite{masnadi2010design}, merging the sampling and learning \cite{hare2011struck}, and employing unlabeled data \cite{grabner2008semi}. One of the most prominent approaches to tackle this problem is to augment the classifier with one or more classifier to break the self-learning loop \cite{tang2007co} and provide a teacher for the main classifier \cite{grabner2008semi}. Such ideas are manifested in the form of co-tracking \cite{tang2007co,meshgi2017efficient} and ensemble tracking \cite{avidan2007ensemble,meshgi2018efficient}.
The bring complementary benefits to the tracker, extra classifier(s) should differ from the main one in training data\cite{avidan2007ensemble}, learning model, update mechanism\cite{zhang2014meem}, update frequency or memory span \cite{meshgi2017efficient}.
Controlling classifier memory is one of the approaches to promote co-tracking and increase accuracy \cite{kalal2012tracking,hare2011struck,zhang2014meem,hong2015multi,wang2016visual,meshgi2017efficient}. 

Here, we take advantage of the information about classifier's uncertainty state in a scenario (Figure \ref{fig:concept}), to control the information exchange between the main classifier of the tracker, and a more accurate yet slow auxiliary classifier in the co-tracking framework. To cope with rapid target changes and handling challenges such as temporal target deformations and occlusion, we set different memory span and update frequency for classifiers. Naturally, the main classifier is selected to an agile, plastic, easily-updatable and frequently-updating model whereas the auxiliary tracker is more sophisticated (accurate yet slow), more stable (memorizing all labels in the tracker's history), and less-frequently updated. The main classifier only queries a label from the auxiliary one, when it is uncertain about a sample's label in line with the uncertainty sampling \cite{lewis1994sequential}. We proposed a Q-learning approach to govern the information exchange between two classifiers w.r.t. the uncertainty state of the first classifier. This scheme automatically balances the stability-plasticity trade-off in tracking \cite{grabner2008semi} and long-short memory trade-off while increasing the speed of the tracker (by avoiding unnecessary queries from the slow classifier) and enhances the generalization ability of the first classifier (by advantaging from the benefits of active learning).   
The proposed tracker performs better than many of the state-of-the-art in tracking-by-detection. 

\section{Proposed Method}
In this section we formalize a tracking-by-detection pipeline, expand it with the notion of traditional and active co-tracking, and elaborate the proposed Q-learning approach that is intended to balance the use of short and long-term memories.

\subsection{Active Co-Tracking}
\label{sec:act}
A tracking-by-detection pipeline consists of a sampler, that selects $n_s$ samples $\mathbf{x}_t=\{x^j_t\}$ from the given frame $I_t$, give it to the classifier modeled by $\theta_t$ to obtain the labels $\mathbf{l}_t=\{\ell^j_t\}$ of the samples. A label is the result of thresholding the scoring function $h:x \rightarrow [0,1]$ for sample $x^j_t$, 
\begin{equation}
\ell^j_t = \mathrm{sign} \Big( h(x^j_t|\theta_t) - \tau \Big)
\label{eq:label}
\end{equation}
and threshold $\tau$ is typically set to \textonehalf. Finally, the model is updated using the obtained labels $\theta_{t+1} = u(\theta_t,\mathbf{x}_t,\mathbf{l}_t)$. 

In an attempt to break the self-learning loop, co-tracking \cite{tang2007co} uses two parallel classifiers $\theta_t^1$ and $\theta_t^2$ with potentially complementary characteristics, and label a sample based on their weighted vote. Both classifiers are updated each frame, and their voting weights $\alpha_t^i$ are re-adjusted based on their label consistency on the co-labeled samples. 
\begin{align}
\ell^j_t &=
  \begin{cases}
   \mathrm{sign} \Big( h(x^j_t|\theta^1_t) - \tau \Big)        &, h(x^j_t|\theta^2_t) < \tau \\
   \mathrm{sign} \Big( h(x^j_t|\theta^2_t) - \tau \Big)        &, h(x^j_t|\theta^1_t) < \tau \\
   \mathrm{sign} \Big(\alpha_t^1 h(x^j_t|\theta^1_t) + \alpha_t^2 h(x^j_t|\theta^2_t) - \tau \Big)         &, \text{otherwise}   
  \end{cases}
  \label{eq:label_co}
\end{align}
Co-tracking is built on the premise that when one classifier has difficulty labeling a sample, the other one assists. Co-tracking increases the tracking accuracy (by decreasing label noise) and addresses model drift.  On the other hand, \textit{(i)} using two classifiers doubles the computational complexity, and \textit{(ii)} in the challenging cases such as temporal full occlusions and background clutter, both of classifiers may encounter difficulty in labeling and will be updated with noisy labels.

To speed up the tracking, Meshgi et al. \cite{meshgi2017active} proposed that a second auxiliary classifier should be consulted, only when needed by the main classifier. This approach enables using more sophisticated models for the auxiliary classifier and decreases computational complexity. To select the samples, that knowing their label maximally help the main classifier to continue accurate tracking, active learning approaches can be used. When a classifier is more uncertain about the label of a sample, knowing its label would increase the learning of the classifier more \cite{lewis1994sequential}, thus improves its convergence speed and generalization with a limited number of labeled samples. In tracking such uncertainty may come from the ineffectiveness of the features, the sample being close to the decision boundary \cite{meshgi2018information}, or missing information (e.g. due to partial occlusion). Therefore, an active co-tracker only queries the labels of the samples from aux classifier when the main classifier is most uncertain about their labels. 
\begin{align}
\ell^j_t &=
  \begin{cases}
   \mathrm{sign} \Big( h(x^j_t|\theta^1_t) - \tau \Big)        &, |h(x^j_t|\theta^1_t) - \tau| > \delta \\
   \mathrm{sign} \Big( h(x^j_t|\theta^2_t) - \tau \Big)        &, \text{otherwise}  
  \end{cases}
  \label{eq:label_act}
\end{align}
Here,  $|h(x^j_t|\theta^1_t) - \tau| > \delta$ means that the uncertainty is less than margin $\delta$.
After the labeling, the main classifier is trained with the co-labeled data and learns maximally from the auxiliary classifier. To handle tracking challenges, such as temporal occlusions and deformations, the aux classifier is updated every $\Delta$ frames. This in-turn further reduces the computational complexity of the tracker and benefits the tracker from a combination of long and short term memories.
\begin{align}
\theta^{(2)}_{t+1} &=
\begin{cases}
 u'(\theta^{(2)}_t,\mathbf{x}_{t-\Delta..t},\mathbf{l}_{t-\Delta..t}) &, t=k\Delta \\
 \theta^{(2)}_t  &, t \neq k\Delta 
\end{cases}
\label{eq:update_long}
\end{align}

The uncertainty margin $\delta$ controls the ``activeness'' of the tracker. Smaller $\delta$ typically reduces the samples that aux classifier labels, focus on recent samples by leaning toward short term memory and reduce the complexity of the tracker. However, such setting put more burden on the main classifier to label the samples correctly and expose the tracker to the model drift, especially if the target observation is noisy or missing (e.g. in case of partial or temporal occlusions). On the contrary, larger $\delta$ increases the number of queries from the aux classifier, exploiting older data at the expense of the speed. 


\subsection{Uncertainty Margin Adjustment via Q-Learning}
\label{sec:qlearning}
Reinforcement learning has been used in visual tracking to learn when to update the tracker \cite{supancic2017tracking}, learn an early decision policy for different frames \cite{zhang2017deep} and to tune tracking hyperparameters \cite{dong2018hyperparameter}.
Here we formulate a Q-learning agent to adjust th euncertainty margin for the active co-tracker.

\begin{figure}[!t]   
\centering
\includegraphics[width= 0.48\linewidth]{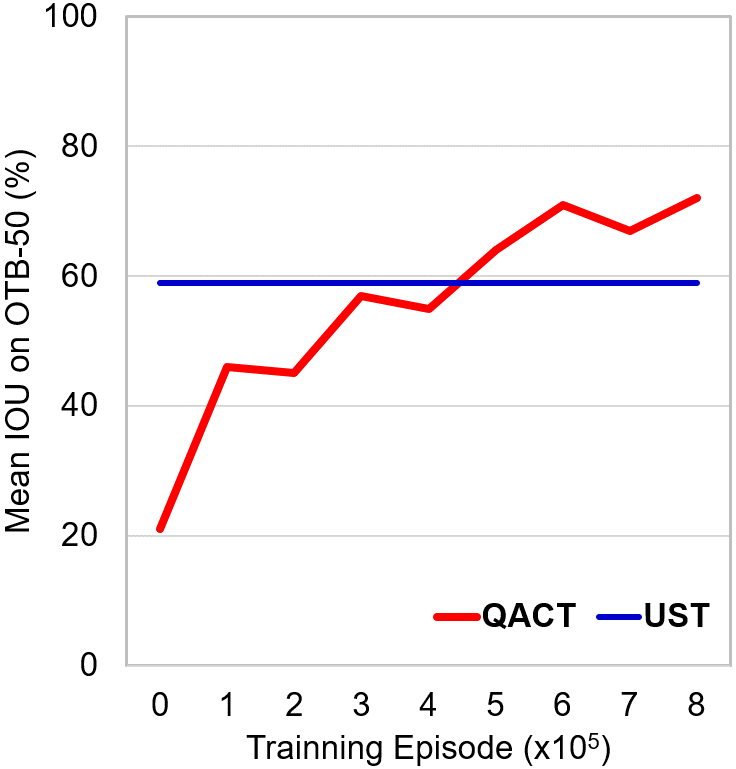}
\includegraphics[width= 0.48\linewidth]{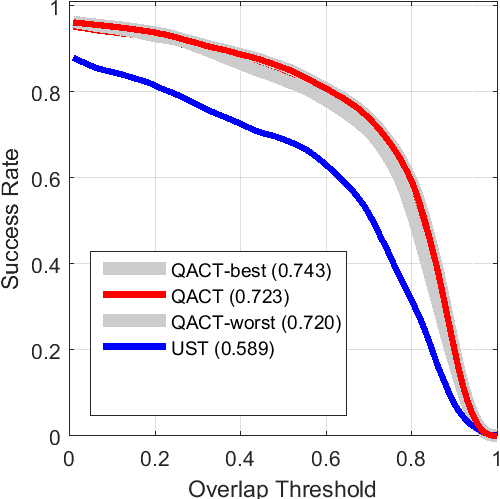}
\caption{The proposed Q-learning agent is trained on YouTubeBB datasets 10 independent runs and their performance on OTB-50 is shaded area (mean performance in red}.
\label{fig:training}
\vspace{-0.7 cm}
\end{figure}

At each time $t$, the agent take an action $a_t$ based on the state $S_t$ of the environment, and the environment gives the reward $r(S_t,a_t)$ and updates its state to $S_{t+1}$. The agent chooses its action w.r.t its policy $\pi(a_t|S_t)$ to maximize the cumulative reward $R_t = \sum_{i=t}^T \gamma^{i-t} r(S_i,a_i)$ where $0<\gamma \leq 1$ is the discount factor to weigh more on earlier rewards. Q-learning proposed to calculate Q-values, the expected maximum scores for each action $a_t$ in state $S_t$, as
\begin{equation}
Q(S_t,a_t) = r(S_t,a_t) + \gamma Q(S_{t+1},a_{t+1})
\end{equation}
\noindent\textbf{State:}
The state $S_t \in \mathbf{S}_t$ of the environment is explained using an $n_b$-bin histogram of uncertainty measurements of main classifier for all samples $\mathbf{x}_t$, 
$$
S_t = f\Big( \mathrm{hist}\big( 1- 2|h(\mathbf{x}_t|\theta^1_t) - \tau| \big) \Big)
$$
To eliminate the effect of the stochastic sampling on the uncertainty histogram, a deterministic sampling approach is used, which obtain $n_s$ equi-distance samples from an area 3x bigger than the target size centered on its last known position. The states are defined by the descriptive features of the histogram $f(.)$, i.e. its mean, variance, and shape. 

\noindent\textbf{Action:} Actions $a_t \in \mathbf{A}_t$ are $n_a$ equi-distanced values in range [0,\textonehalf] to be assigned to margin $\delta$. 

\noindent\textbf{Reward:}
During training time, the reward is defined as the intersection-over-union (IOU) of the target estimation and ground truth and if the IOU exceeds 90\%, the reward triples. Contrarily, if the IOU drops under 50\%, no reward is given to the learner, and if remains under 50\% for five consecutive frames, the negative reward of -3 is given to the learner.

\subsection{Q-learning for Active Co-Tracking}

In the proposed QACT tracker (Figure \ref{fig:schematic}), the main classifier is a KNN classifier, with color names and HOG features. In each frame, $t$, $n_s$ co-labeled samples are added and samples older than $t-\Delta$ are discarded from KNN. The aux classifier is a part-based SVM classifier \cite{felzenszwalb2010object}, that is retrained every $\Delta$ frame with all of the co-labeled samples from the beginning of the tracking. In each frame, the target location and scale is defined with a weighted vote of the obtained positive samples, by considering classification score as their weight. 

The parameters of the tracker, except those related to the Q-learning module, is defined using cross-validation on the OTB-50 dataset \cite{wu2013online}. The Q-learning parameters are then set to fixed values of $n_b=100$, $n_a=25$, and $\gamma = 0.99$ and the Q-values are randomly initialized with a small positive white noise.
The proposed tracker is trained on annotated frames of YouTubeBB dataset \cite{real2017youtube} to train the proposed Q-Learning method using the Boltzmann-Gumbel exploration \cite{cesa2017boltzmann} to exploit all the information present in the estimated Q-values with an annealed temperature parameter. For each of 800,000 training episode, we randomly sample a 20-seconds long clip with one annotation for each one second, and provide zero reward for agent in between annotations. During run-time, the tracker greedily selects the action $a^*_t$ which yields the highest expected reward, $a^*_t = \argmax_{a'_t \in \mathbf{A}_t} Q(S_t, a'_t)$.

\section{Evaluation}
\label{sec3}

\begin{figure}[!t]   
\centering
\includegraphics[width= 0.49\linewidth]{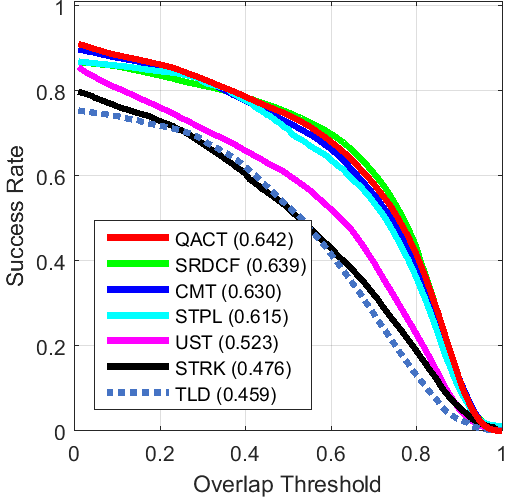}
\includegraphics[width= 0.49\linewidth]{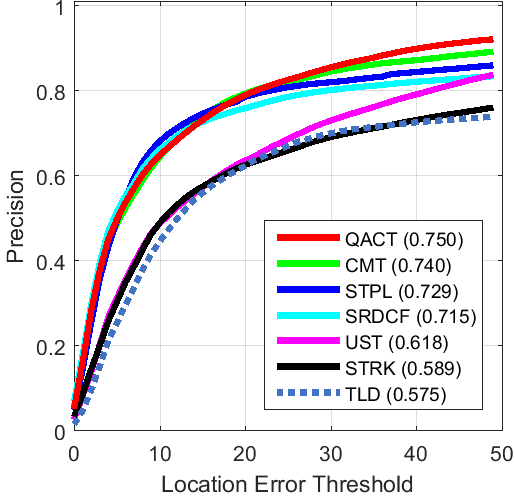}
\caption{Quantitative evaluation of trackers using precision plot (left) and success plot (right) for all videos in OTB-100\cite{wu2015object}.}
\label{fig:precision}
\vspace{-0.3 cm}
\end{figure}
\begin{figure}[!t]
\centering
\includegraphics[width= 0.32\linewidth]{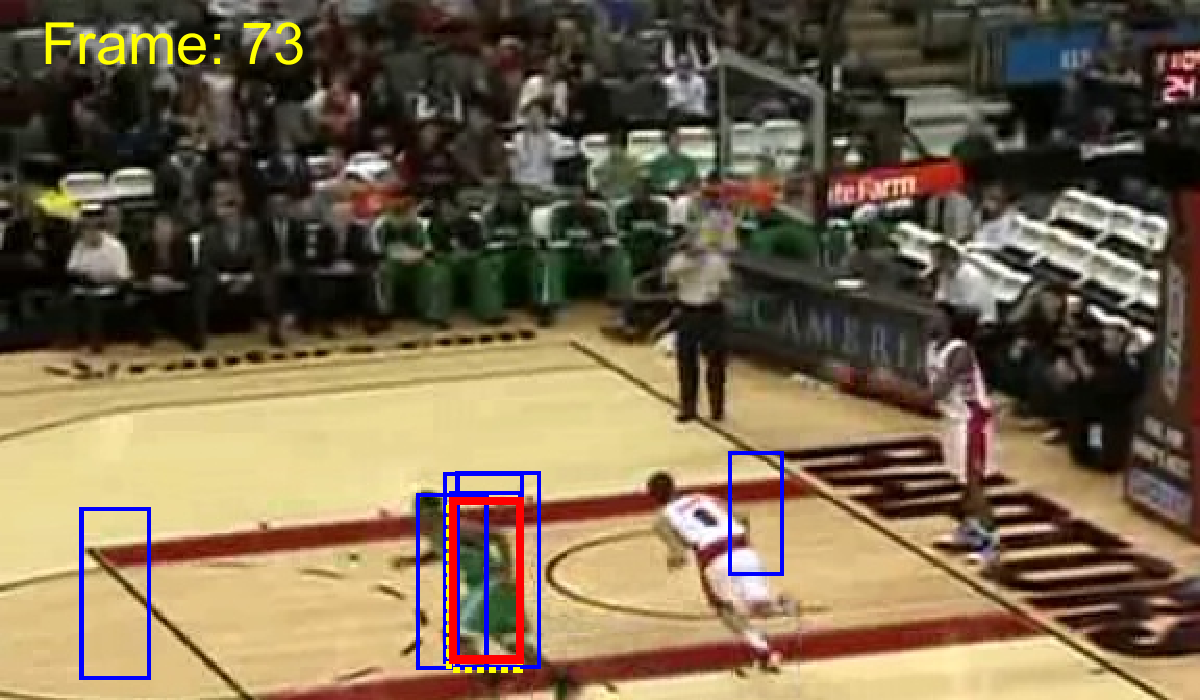}
\includegraphics[width= 0.32\linewidth]{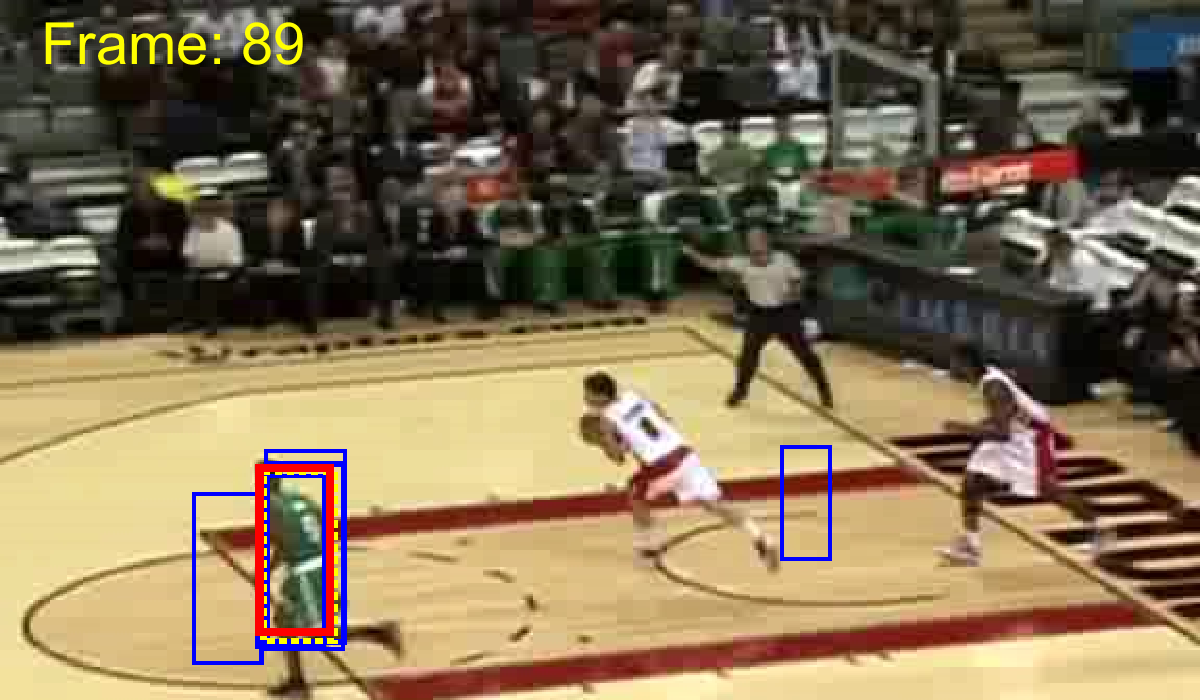}
\includegraphics[width= 0.32\linewidth]{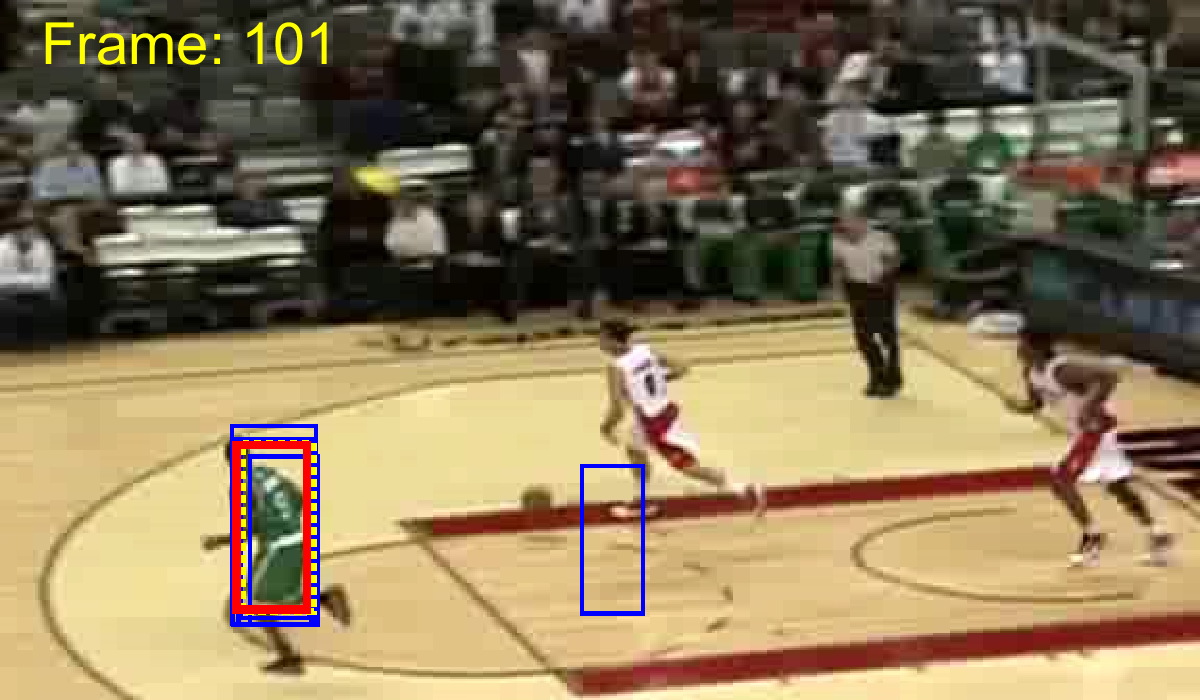}
\includegraphics[width= 0.32\linewidth]{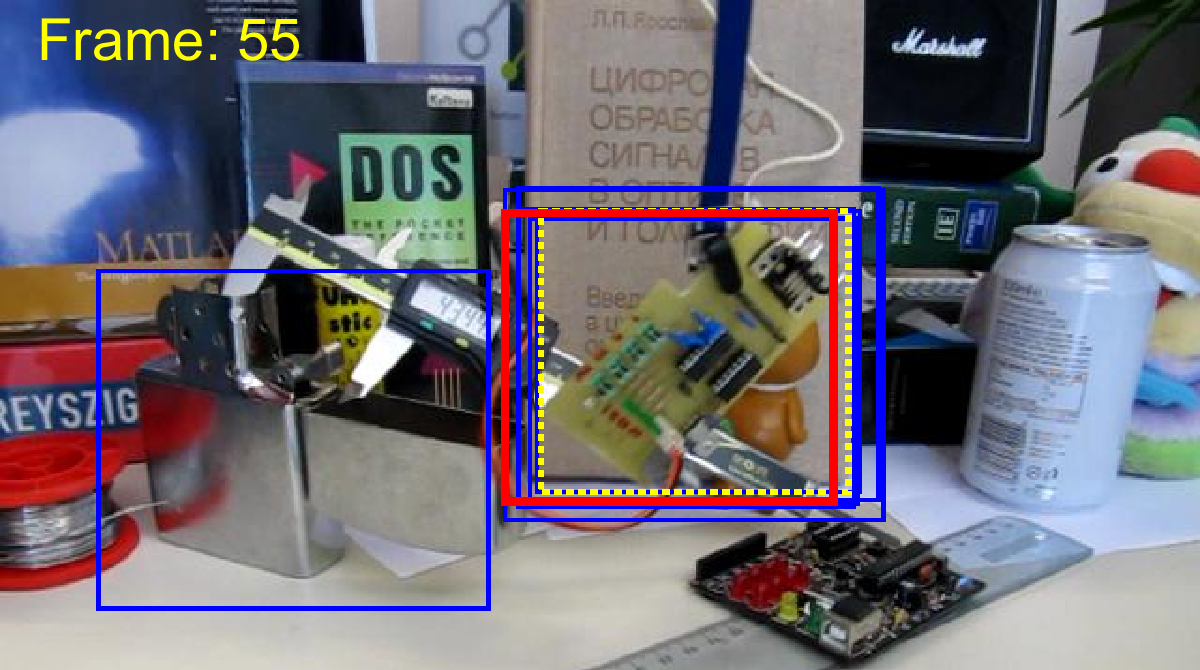}
\includegraphics[width= 0.32\linewidth]{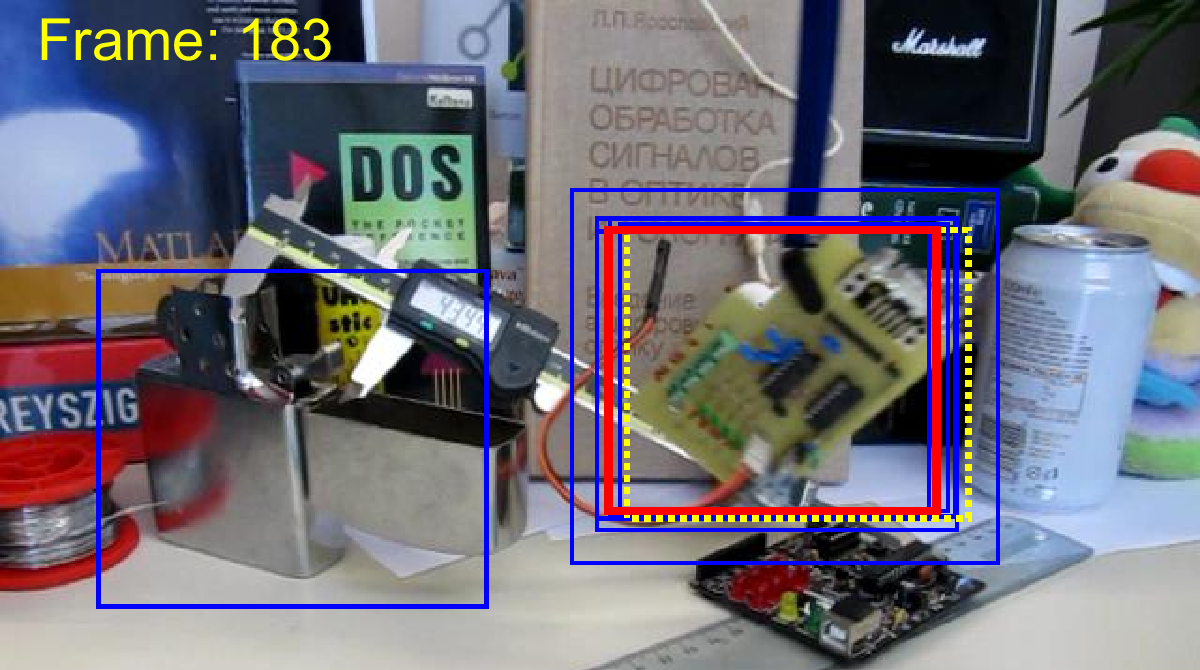}
\includegraphics[width= 0.32\linewidth]{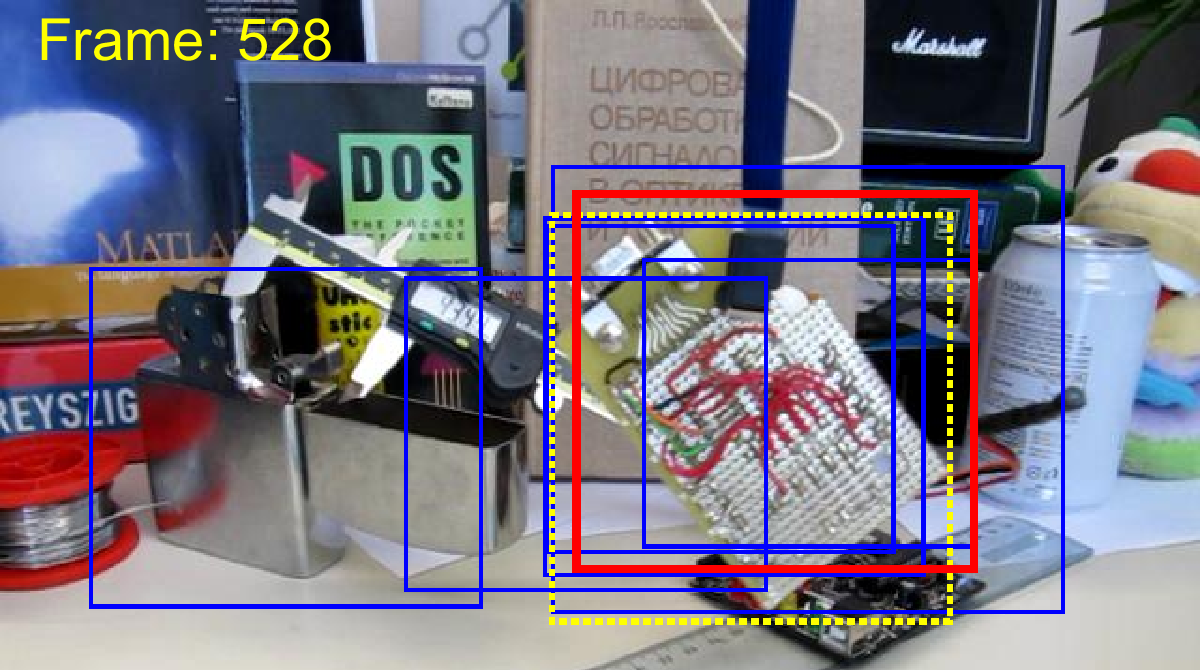}
\includegraphics[width= 0.32\linewidth]{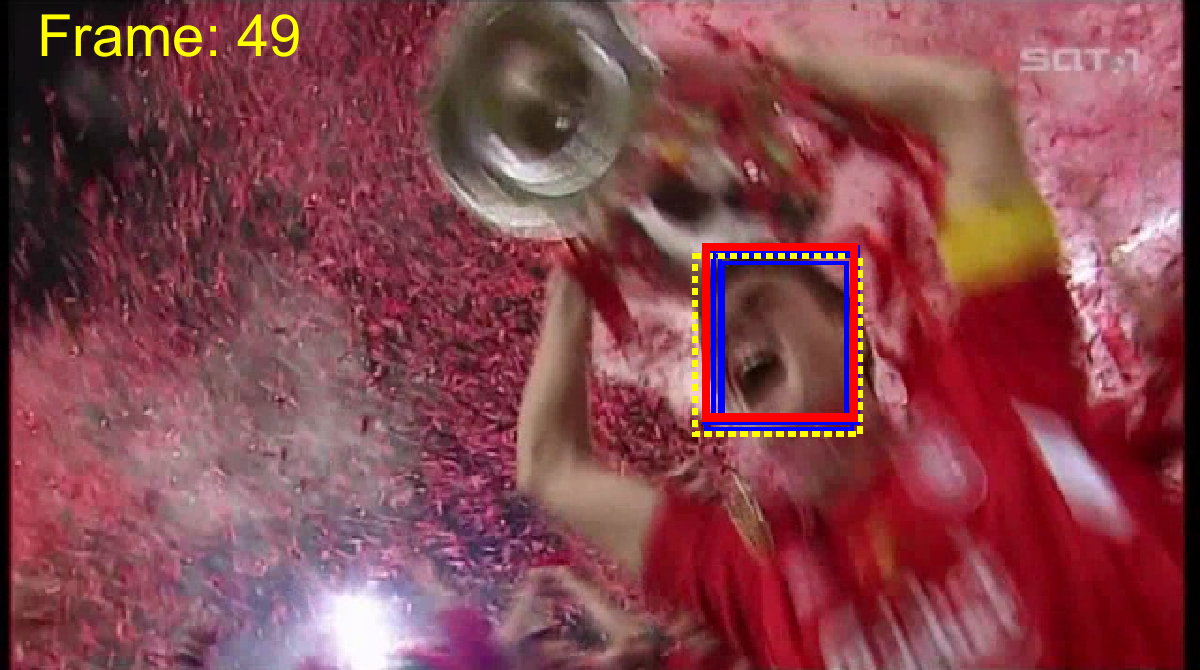}
\includegraphics[width= 0.32\linewidth]{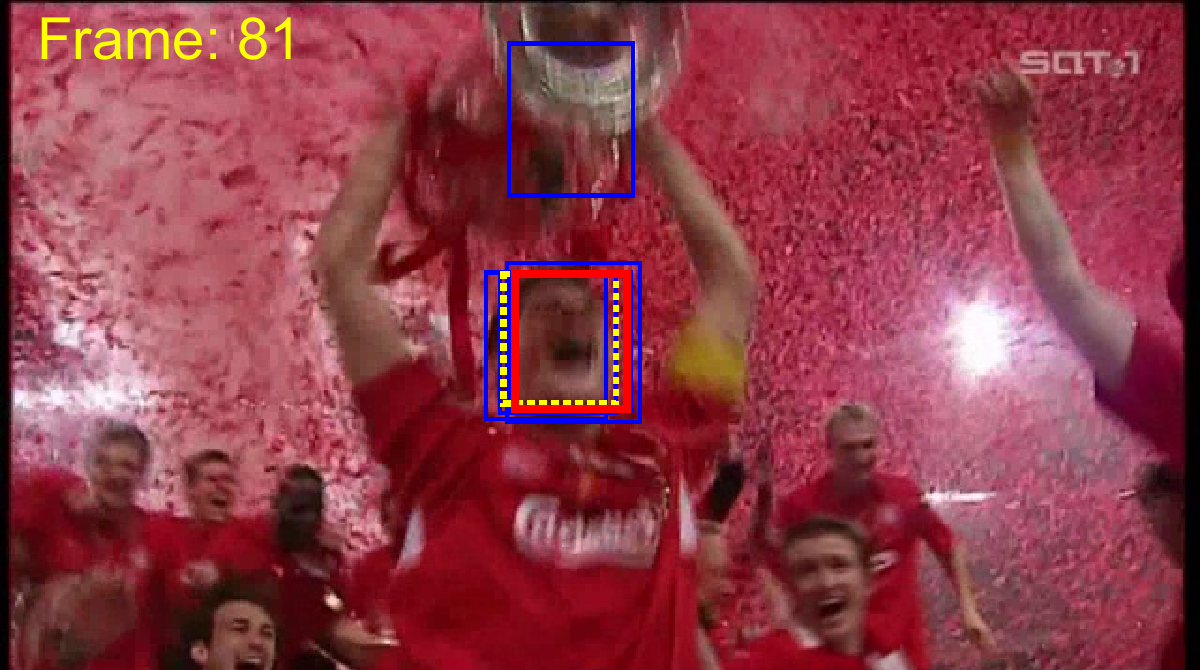}
\includegraphics[width= 0.32\linewidth]{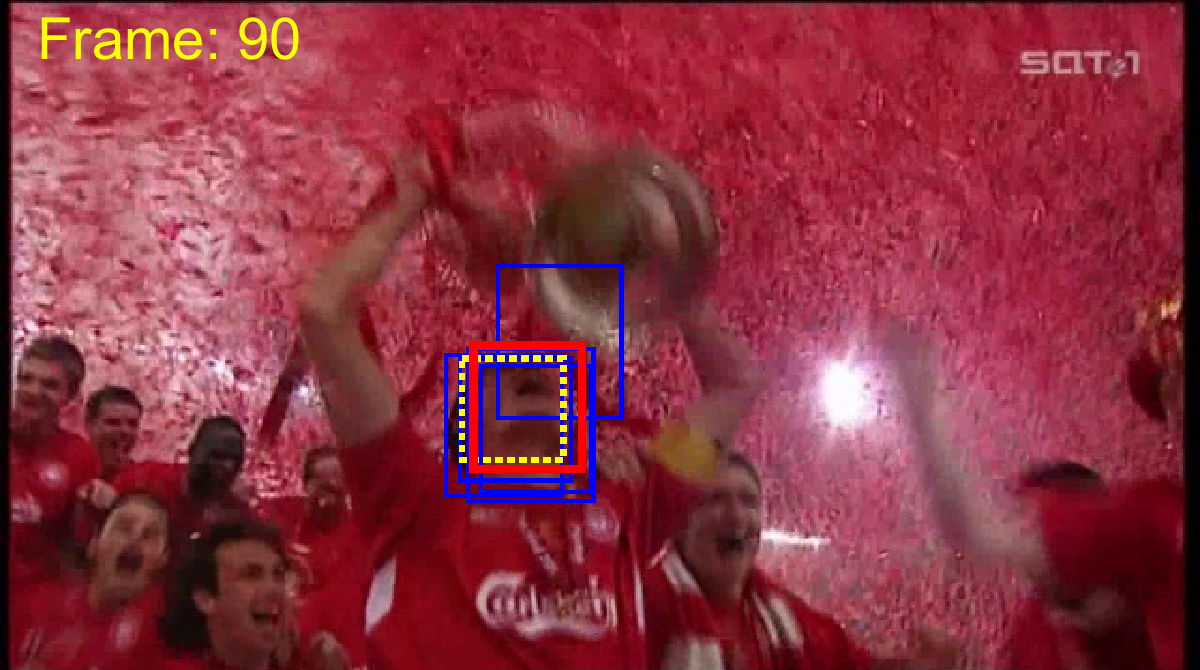}
\caption{Exemplary tracking results of proposed tracker (in red) and other evaluated trackers (blue) on several challenging video sequences. The ground truth is depicted in yellow. More results are available from \texttt{author's page}.
}
\label{fig:eval_qual}
\vspace{-0.5cm}
\end{figure}

\vspace*{-\baselineskip}
\begin{table}[!b]
\caption{Quantitative evaluation of trackers under different tracking challenges using AUC(\%) of success plot on OTB-50. The {\color{red}first}, {\color{green}second} and {\color{blue}third} best results are highlighted. Scenario attributes indicate changes in illumination, scale, in-plane and out-of-plane rotation, deformation, occlusion, out-of-view, clutter, low resolution, fast motion and motion blur.}
\label{tab:attributes}
\centering
\scalebox{0.85}{
\renewcommand{\arraystretch}{0.9}
\begin{tabular}{@{}>{\bfseries}l c c c c c c c c@{}} \toprule
\makebox[9mm]{Attribute}& \makebox[7mm]{KNN+} &\makebox[7mm]{SVM+}  &\makebox[7mm]{TLD }  &\makebox[7mm]{STRK}  &\makebox[7mm]{UST}  &\makebox[7mm]{MEEM}  &\makebox[7mm]{MSTR}   & \makebox[7mm]{QACT}\\ \midrule
    IV                  & {\color{black}24.1} & {\color{black}39.9} & {\color{black}47.8} & {\color{black}53.0} & {\color{black}58.5} & {\color{blue} 62.3} & {\color{red}  72.6} & {\color{red}  72.6} \\ 
    SV                  & {\color{black}23.0} & {\color{black}42.4} & {\color{black}49.1} & {\color{black}50.7} & {\color{blue} 58.8} & {\color{black}58.3} & {\color{green}70.6} & {\color{red}  72.3} \\  
    IPR                 & {\color{black}25.3} & {\color{black}44.4} & {\color{black}50.4} & {\color{black}53.7} & {\color{blue} 61.9} & {\color{black}57.7} & {\color{green}68.5} & {\color{red}  73.4} \\  
    OPR                 & {\color{black}25.8} & {\color{black}43.1} & {\color{black}47.8} & {\color{black}53.2} & {\color{black}59.7} & {\color{blue} 62.1} & {\color{green}70.2} & {\color{red}  70.4} \\  
    DEF                 & {\color{black}28.9} & {\color{black}41.0} & {\color{black}38.2} & {\color{black}51.3} & {\color{black}55.9} & {\color{blue} 61.9} & {\color{red}  68.9} & {\color{green}66.1} \\  
    OCC                 & {\color{black}23.5} & {\color{black}39.9} & {\color{black}46.1} & {\color{black}50.2} & {\color{black}58.6} & {\color{blue} 60.8} & {\color{green}71.0} & {\color{red}  71.7} \\  
    OV                  & {\color{black}27.7} & {\color{black}52.0} & {\color{black}53.5} & {\color{black}51.5} & {\color{black}56.9} & {\color{blue} 68.5} & {\color{red}  73.3} & {\color{green}71.1} \\  
    LR                  & {\color{black}13.3} & {\color{black}13.6} & {\color{black}36.2} & {\color{black}33.3} & {\color{black}33.1} & {\color{blue} 43.5} & {\color{green}50.2} & {\color{red}  56.0} \\  
    BC                  & {\color{black}30.7} & {\color{black}40.0} & {\color{black}39.4} & {\color{black}51.5} & {\color{black}48.0} & {\color{blue} 67.0} & {\color{red}  71.7} & {\color{green}71.1} \\  
    FM                  & {\color{black}23.0} & {\color{black}43.2} & {\color{black}44.6} & {\color{black}52.0} & {\color{black}53.4} & {\color{green}64.6} & {\color{red}  65.0} & {\color{blue} 64.3} \\  
    MB                  & {\color{black}22.9} & {\color{black}35.0} & {\color{black}41.0} & {\color{black}46.7} & {\color{black}45.2} & {\color{blue} 62.8} & {\color{green}65.2} & {\color{red}  65.6} \\ \midrule

    ALL                    & {\color{black}27.8} & {\color{black}43.5} & {\color{black}49.3} & {\color{black}54.8} & {\color{black}58.9} & {\color{blue} 61.7} & {\color{green}71.8} & {\color{red}  72.3} \\ \midrule
    FPS                 & {\color{red}  76.6} & {\color{black} 3.8} & {\color{black}21.2} & {\color{black}11.3} & {\color{green}28.3} & {\color{black}14.2} & {\color{black} 8.3} & {\color{blue} 27.1} \\ \bottomrule
\end{tabular}
}
\vspace{-0.5cm}
\end{table}
The proposed tracker is evaluated against its baselines, competitive trackers which leverage memory control for tracking, and finally the state-of-the-art in tracking-by-detection. The experiments are conducted using OTB-50 \cite{wu2013online} and OTB-100 \cite{wu2015object} benchmarks. Success and precision plots are used to compare the performance of the trackers.

Figure \ref{fig:training} shows that the proposed tracker (QACT) outperforms its baseline, UST (uncertainty sampling tracker) using Q-learning to tune the uncertainty margin parameter. The shaded area indicates 10 independent training runs for the Q-learning method, and the red plot indicates the mean of these runs and will be used hereafter as the QACT result to be compared with other algorithms. 
Table \ref{tab:attributes} contains the ablation study as well as a comparison with TLD\cite{kalal2012tracking}, STRK\cite{hare2011struck}, MEEM\cite{zhang2014meem} and MSTR\cite{hong2015multi} that handles the memory of the tracker to improve tracking. UST, is the active co-tracker with dual memory explained in section \ref{sec:act}, with KNN+\cite{meshgi2017efficient} as short-term classifier and SVM+ as the long-term one. As the table shows, QACT demonstrates superior performance on many aspects of tracking compared to its baselines and to other competitive methods. It is interesting to note the low-resolution case (LR) that the active co-tracker made from very low-performance trackers (i.e. UST), has a significantly higher performance from its classifiers (KNN+ and SVM+), still by online adjustment of uncertainty margin the performance of QACT has another large improvement. Finally, Figure \ref{fig:precision} depicts the promising performance of  proposed approach compared to the state-of-the-art in tracking by detection such as CMT \cite{meshgi2017active}, STPL \cite{bertinetto2016staple} and SRDCF \cite{danelljan2015learning}.

\section{Conclusion}
\label{sec:conclusion}
Active co-tracking combines the results of two classifiers, by using one of them on-demand triggered by the uncertainty of the other classifier. The threshold to trigger is of paramount importance to balance the information exchange between two classifiers, especially if they differ in accuracy, speed, model update frequency, and retraining complexity. We proposed a Q-learning mechanism that monitors the uncertainty state of the first classifiers to control the trigger. We elaborated the active co-tracking framework with dual memory, the design and training of the Q-learning for parameter tuning, and presented the superior performance of this tracker.

\vfill
\pagebreak


\bibliographystyle{IEEEbib}
\def\IEEEbibitemsep{-2pt plus .5pt}
\bibliography{refs}

\begin{thebibliography}{10}

\bibitem{masnadi2010design}
Hamed Masnadi-Shirazi, Vijay Mahadevan, and Nuno Vasconcelos,
\newblock ``On the design of robust classifiers for computer vision,''
\newblock in {\em CVPR'10}, 2010.

\bibitem{hare2011struck}
Sam Hare, Amir Saffari, and Philip~HS Torr,
\newblock ``Struck: Structured output tracking with kernels,''
\newblock in {\em ICCV'11}, 2011.

\bibitem{grabner2008semi}
Helmut Grabner, Christian Leistner, and Horst Bischof,
\newblock ``Semi-supervised on-line boosting for robust tracking,''
\newblock in {\em ECCV'08}. 2008.

\bibitem{tang2007co}
Feng Tang, Shane Brennan, Qi~Zhao, and Hai Tao,
\newblock ``Co-tracking using semi-supervised support vector machines,''
\newblock in {\em ICCV'07}.

\bibitem{meshgi2017efficient}
Kourosh Meshgi, Mirzaei~Maryam Sadat, Shigeyuki Oba, and Shin Ishii,
\newblock ``Efficient asymmetric co-tracking using uncertainty sampling,''
\newblock in {\em IEEE ICSIPA'17}.

\bibitem{avidan2007ensemble}
Shai Avidan,
\newblock ``Ensemble tracking,''
\newblock {\em PAMI}, vol. 29, 2007.

\bibitem{meshgi2018efficient}
Kourosh Meshgi, Shigeyuki Oba, and Shin Ishii,
\newblock ``Efficient diverse ensemble for discriminative co-tracking,''
\newblock in {\em CVPR'18}.

\bibitem{zhang2014meem}
Jianming Zhang, Shugao Ma, and Stan Sclaroff,
\newblock ``Meem: Robust tracking via multiple experts using entropy
  minimization,''
\newblock in {\em ECCV'14}.

\bibitem{kalal2012tracking}
Zdenek Kalal, Krystian Mikolajczyk, and Jiri Matas,
\newblock ``Tracking-learning-detection,''
\newblock {\em PAMI}, vol. 34, no. 7, pp. 1409--1422, 2012.

\bibitem{hong2015multi}
Zhibin Hong, Zhe Chen, Chaohui Wang, Xue Mei, Danil Prokhorov, and Dacheng Tao,
\newblock ``Multi-store tracker (muster): a cognitive psychology inspired
  approach to object tracking,''
\newblock in {\em CVPR'15}.

\bibitem{wang2016visual}
Shu Wang, Shaoting Zhang, Wei Liu, and Dimitris~N Metaxas,
\newblock ``Visual tracking with reliable memories.,''
\newblock in {\em IJCAI}, 2016, pp. 3491--3497.

\bibitem{lewis1994sequential}
David~D Lewis and William~A Gale,
\newblock ``A sequential algorithm for training text classifiers,''
\newblock in {\em ACM SIGIR'94}, 1994, pp. 3--12.

\bibitem{meshgi2017active}
Kourosh Meshgi, Shigeyuki Oba, and Shin Ishii,
\newblock ``Active discriminative tracking using collective memory,''
\newblock in {\em MVA'17}.

\bibitem{meshgi2018information}
Kourosh Meshgi, Maryam~Sadat Mirzaei, and Shigeyuki Oba,
\newblock ``Information-maximizing sampling to promote tracking-by-detection,''
\newblock in {\em ICIP'18}, 2018.

\bibitem{supancic2017tracking}
James Supancic~III and Deva Ramanan,
\newblock ``Tracking as online decision-making: Learning a policy from
  streaming videos with reinforcement learning,''
\newblock in {\em ICCV'17}.

\bibitem{zhang2017deep}
Da~Zhang, Hamid Maei, Xin Wang, and Yuan-Fang Wang,
\newblock ``Deep reinforcement learning for visual object tracking in videos,''
\newblock {\em arXiv preprint arXiv:1701.08936}, 2017.

\bibitem{dong2018hyperparameter}
Xingping Dong, Jianbing Shen, Wenguan Wang, Yu~Liu, Ling Shao, and Fatih
  Porikli,
\newblock ``Hyperparameter optimization for tracking with continuous deep
  q-learning,''
\newblock in {\em CVPR'18}.

\bibitem{cesa2017boltzmann}
Nicol{\`o} Cesa-Bianchi, Claudio Gentile, G{\'a}bor Lugosi, and Gergely Neu,
\newblock ``Boltzmann exploration done right,''
\newblock in {\em NIPS'17}, 2017, pp. 6284--6293.

\bibitem{van2009learning}
Joost Van De~Weijer, Cordelia Schmid, Jakob Verbeek, and Diane Larlus,
\newblock ``Learning color names for real-world applications,''
\newblock {\em IEEE TIP}, vol. 18, no. 7, pp. 1512--1523, 2009.

\bibitem{dalal2005histograms}
Navneet Dalal and Bill Triggs,
\newblock ``Histograms of oriented gradients for human detection,''
\newblock in {\em Computer Vision and Pattern Recognition, 2005. CVPR 2005.
  IEEE Computer Society Conference on}. IEEE, 2005, vol.~1, pp. 886--893.

\bibitem{felzenszwalb2010object}
Pedro~F Felzenszwalb, Ross~B Girshick, David McAllester, and Deva Ramanan,
\newblock ``Object detection with discriminatively trained part-based models,''
\newblock {\em PAMI}, vol. 32, no. 9, pp. 1627--1645, 2010.

\bibitem{wu2013online}
Yi~Wu, Jongwoo Lim, and Ming-Hsuan Yang,
\newblock ``Online object tracking: A benchmark,''
\newblock in {\em CVPR'13}. IEEE, 2013, pp. 2411--2418.

\bibitem{real2017youtube}
Esteban Real, Jonathon Shlens, Stefano Mazzocchi, Xin Pan, and Vincent
  Vanhoucke,
\newblock ``Youtube-boundingboxes: A large high-precision human-annotated data
  set for object detection in video,''
\newblock in {\em CVPR'17}. IEEE, 2017, pp. 7464--7473.

\bibitem{wu2015object}
Yi~Wu, Jongwoo Lim, and Ming-Hsuan Yang,
\newblock ``Object tracking benchmark,''
\newblock {\em PAMI}, 2015.

\bibitem{bertinetto2016staple}
Luca Bertinetto, Jack Valmadre, Stuart Golodetz, Ondrej Miksik, and Philip~HS
  Torr,
\newblock ``Staple: Complementary learners for real-time tracking,''
\newblock in {\em Proceedings of the IEEE Conference on Computer Vision and
  Pattern Recognition}, 2016, pp. 1401--1409.

\bibitem{danelljan2015learning}
Martin Danelljan, Gustav Hager, Fahad Shahbaz~Khan, and Michael Felsberg,
\newblock ``Learning spatially regularized correlation filters for visual
  tracking,''
\newblock in {\em ICCV'15}, 2015, pp. 4310--4318.

\end{thebibliography}

\end{document}